\documentclass[11pt]{article}

\usepackage[final]{acl}

\usepackage{times}
\usepackage{latexsym}

\usepackage[T1]{fontenc}

\usepackage[utf8]{inputenc}

\usepackage{microtype}

\usepackage{inconsolata}

\usepackage{graphicx}
\usepackage{booktabs}
\usepackage{multirow}
\usepackage{amsmath}
\usepackage{amssymb}
\usepackage{array}
\usepackage{tcolorbox}
\tcbset{
  colback=gray!3,
  colframe=gray!50,
  boxrule=0.5pt,
  arc=1mm,
  left=1mm,
  right=1mm,
  top=1mm,
  bottom=1mm
}
\title{PromptRad: Knowledge-Enhanced Multi-Label Prompt-Tuning for Low-Resource Radiology Report Labeling}

\author{
 \textbf{Ying-Jia Lin\textsuperscript{1}},
 \textbf{Tzu-Chin Lo\textsuperscript{2}}\thanks{Work was done while the author was affiliated with Chang Gung Memorial Hospital.},
 \textbf{Ping-Chien Li\textsuperscript{3}},
 \textbf{Chi-Tung Cheng\textsuperscript{4}},
\\
 \textbf{Chien-Hung Liao\textsuperscript{4}},
 \textbf{Hung-Yu Kao\textsuperscript{5}}
\\
 \textsuperscript{1}Department of Artificial Intelligence and AI Research Center, Chang Gung University,\\
 \textsuperscript{2}Department of Radiology, Sijhih Cathay General Hospital,\\
 \textsuperscript{3}Department of Medical Imaging and Intervention, Chang Gung Memorial Hospital,\\
 \textsuperscript{4}Department of Trauma and Emergency Surgery, Chang Gung Memorial Hospital, \\
 \textsuperscript{5}Department of Computer Science, National Tsing Hua University\\
 \texttt{yjlin@cgu.edu.tw, hykao@cs.nthu.edu.tw} \\
 }

\begin{document}
\maketitle
\begin{abstract}
Automatic report labeling facilitates the identification of clinical findings from unstructured text and enables large-scale annotation for medical imaging research.
Existing rule-based labelers struggle with the diverse descriptions in clinical reports, while fine-tuning pre-trained language models (PLMs) requires large amounts of labeled data that are often unavailable in clinical settings.
In this paper, we propose PromptRad, a knowledge-enhanced multi-label \textbf{prompt}-tuning approach for \textbf{rad}iology report labeling under low-resource settings.
PromptRad reformulates multi-label classification as masked language modeling and incorporates synonyms from the UMLS Metathesaurus into a multi-word verbalizer to enrich category representations.
By fine-tuning the PLM without additional classification layers, PromptRad requires substantially less labeled data than conventional
fine-tuning. Experiments on liver CT (computed tomography) reports show that PromptRad outperforms dictionary-based and fine-tuning baselines with only 32 labeled training examples, and achieves competitive performance with GPT-4 despite using a much smaller model.
Further analysis demonstrates that PromptRad captures complex negation patterns more effectively than existing methods, making it a promising solution for report labeling in data-scarce clinical scenarios.
Our code is available at \url{https://github.com/ila-lab/PromptRad}.
\end{abstract}

\section{Introduction}
Radiology reports contain valuable medical findings about patients' conditions and play a crucial role in clinical decision-making.
However, these reports are usually lengthy and unstructured (see Figure~\ref{fig:example_report} for an example), making it difficult to efficiently extract clinical information \cite{chen2017,pons2016}.
Traditional approaches rely on expert knowledge from the Unified Medical Language System (UMLS)~\cite{umls2004} and computational linguistics techniques for medical concept extraction \cite{aronson2001effective, Wang_2017_CVPR, peng2018negbio}.
However, these dictionary-based methods depend on pre-established mappings between text and medical concepts and often fail when reports use diverse or non-conventional descriptions \cite{irvin2019chexpert}.
While pre-trained language models (PLMs) such as BioBERT \cite{biobert} and PubMedBERT \cite{pubmedbert} have been applied to report labeling \cite{pmlr-v121-wood20a, smit-etal-2020-combining, Li2022}, fine-tuning these models requires substantial labeled data \cite{devlin2019bert, dodge2020, zhang2021revisiting}, which is scarce in clinical settings due to the need for domain expertise during annotation.

\begin{figure}[t]
  \centering
  \small
  \fbox{\parbox{0.95\columnwidth}{%
    \textbf{Example Liver CT Report (de-identified):}\\[4pt]
    CT study of chest and abdomen without and with IV contrast
    enhancement shows:\\[2pt]
    1. Rapid enlarging nodule or mass at RUL, suggestive of rapid
    lung \underline{metastasis}.\\
    2. Multiple new small nodules (0.2--1.0 cm) at bilateral lungs.\\
    3. A 4cm liver mass at S5, with peripheral globular enhancement,
    favor \underline{hemangioma}.\\
    4. Small \underline{cysts} at right lobe of liver.\\
    5. Gallstones. \\[4pt]
    \textbf{Gold Labels}: Cyst~\textcolor{green!60!black}{\checkmark},
    Metastasis~\textcolor{green!60!black}{\checkmark},
    Hemangioma~\textcolor{green!60!black}{\checkmark}
  }}
  \caption{An example liver CT report from our dataset with annotated findings. Underlined terms indicate relevant mentions for each positive label.}
  \label{fig:example_report}
\end{figure}

Prompt-tuning \cite{gao-etal-2021-making, pet2021} offers a promising alternative by transforming classification into masked language modeling, enabling PLMs to perform well with limited labeled data \cite{schick-schutze-2021-just}.
However, existing prompt-tuning methods \cite{hu-etal-2022-knowledgeable, gao-etal-2021-making, 10.1145/3560815} are designed for multi-class classification with mutually exclusive categories. In
contrast, radiology report labeling is inherently a multi-label task, where multiple findings may coexist in a single report.
This limits the direct applicability of existing prompt-tuning approaches to report labeling.

We propose PromptRad, a knowledge-enhanced multi-label \textbf{prompt}-tuning approach for \textbf{rad}iology report labeling under low-resource settings.
Here, ``low-resource'' refers to settings where only a limited number of expert-annotated reports are available for training.
PromptRad adapts a pre-trained masked language model \cite{devlin2019bert, pubmedbert} for multi-label classification through prompt-tuning.
Inspired by Knowledgeable Prompt-Tuning (KPT)~\cite{hu-etal-2022-knowledgeable}, we design a multi-word verbalizer that incorporates synonyms from the UMLS Metathesaurus \cite{umls2004} as label-to-word mappings, enriching category representations for clinical contexts.
We further develop automatic prompt generation based on \citet{gao-etal-2021-making} to explore the space of textual templates and enhance performance.

Experiments on liver CT reports from a large medical center demonstrate that PromptRad outperforms dictionary-based methods \cite{aronson2001effective, peng2018negbio} and fine-tuning baselines, and is competitive with GPT-4 \cite{openai2023gpt} even with only 32 labeled training examples.
Our analysis of negation cases shows that PromptRad captures complex negation patterns more effectively than rule-based approaches, demonstrating its robustness for clinical report labeling.

In summary, our contributions are threefold:
\begin{itemize}
    \item We formulate low-resource radiology report labeling as a multi-label prompt-tuning problem and adapt a masked language model \cite{devlin2019bert} to predict multiple clinical findings from a single report.
    \item We introduce a UMLS-informed multi-word verbalizer and an automatic prompt generation strategy to improve label representation and template selection in clinical contexts.
    \item We evaluate PromptRad on a real-world liver CT report dataset and analyze its behavior under limited supervision, including challenging negation cases.
\end{itemize}

\section{Related Work}\label{sec:related_work}
\subsection{Report Labeling}
Traditional approaches to report labeling rely on pre-built knowledge bases.
MetaMap~\cite{aronson2001effective} maps medical text to UMLS concepts, and NegBio~\cite{peng2018negbio} extends it with dependency parsing rules for negation detection. CheXpert~\cite{irvin2019chexpert} replaces predefined concept extractors with manually curated mention lists and more sophisticated negation rules for chest X-ray (CXR) reports.
However, these rule-based methods depend on predefined patterns and may fail to generalize across report types and institutions.

Machine learning approaches, including CNN-based \cite{majkowska2020chest,shin2017cnn} and LSTM-based methods \cite{Dahl2021,bmcmd22}, have also been explored for report labeling but require large amounts of labeled data.
More recently, PLM-based approaches have shown strong performance: ALARM~\cite{pmlr-v121-wood20a} uses BioBERT~\cite{biobert} for MRI reports, CheXbert~\cite{smit-etal-2020-combining} uses BlueBERT~\cite{peng-etal-2019-transfer} for CXR reports, and other studies have applied BERT~\cite{devlin2019bert,Li2022} and domain-specific PLMs~\cite{nowak2023transformer} to various report labeling tasks.
While effective, fine-tuning PLMs typically requires substantial labeled data~\cite{NEURIPS2020_1457c0d6,dodge2020,zhang2021revisiting}, which is scarce in clinical settings.
Unlike most prior radiology labelers that focus on chest X-ray reports or rely on large annotated corpora, our work focuses on low-resource multi-label labeling for liver CT reports.

\subsection{Prompt-Tuning}
Prompt-tuning~\cite{pet2021,gao-etal-2021-making} reformulates classification as masked language modeling, reducing reliance on newly initialized, task-specific classification layers and enabling effective few-shot learning~\cite{schick-schutze-2021-just}.
Prompt-based methods have also been explored in biomedical NLP tasks, including biomedical relation extraction~\cite{he_prompt_2024} and lay summary generation~\cite{wu-etal-2023-ikm}.
KPT~\cite{hu-etal-2022-knowledgeable} improves prompt-tuning by incorporating external knowledge into the verbalizer.
\citet{sportprompt2022} extended prompt-tuning to multi-label text classification and proposed PTMLTC (Prompt Tuning Method for Multi-Label Text Classification), which is the closest to our work.
However, PTMLTC was designed for the educational domain and relies on a single-word verbalizer, which may not capture the diverse expressions for the target categories.
To the best of our knowledge, no prior work has applied prompt-tuning to multi-label radiology report labeling under low-resource settings, especially using small local models with few-shot training.

\subsection{Recent Advances in LLM-based Report Labeling}
With the emergence of large language models (LLMs), recent work has explored LLM-based approaches for radiology report analysis.
CheX-GPT~\cite{gu2024chexgpt} uses GPT-4 as a zero-shot labeler and then trains a BERT-based model on 50k pseudo-labeled chest X-ray reports generated by GPT-4.
\citet{fytas-etal-2024-radprompt} propose a multi-turn prompting strategy that combines rule-based insights with LLM predictions for chest X-ray report classification.
Domain-adapted models such as Radiology-Llama2~\cite{liu2023radiologyllama2} fine-tune open-source LLMs on radiology corpora via instruction tuning, and
\citet{abdullah2025llmreportlabeling} fine-tune a smaller LLM for report labeling on the MIMIC-CXR dataset.
However, these approaches either require large volumes of (pseudo-)labeled data, domain-specific corpora for continued pre-training, or access to proprietary APIs, raising concerns regarding deployment cost and data privacy in clinical settings.
In contrast, PromptRad is a prompt-tuning approach that fine-tunes a masked language model with a UMLS-informed verbalizer, requiring only 32 labeled examples and no access to external APIs.

\begin{table}[htbp]
  \centering
  \caption{Number of positive findings per category in our labeled clinical report dataset. HCC: Hepatocellular Carcinoma.}
  \label{tab:dataset}
  \begin{tabular}{@{}lcc@{}}
    \toprule
    \multirow{2}{*}{Category} & Training & Test \\
                               & (2008--2014) & (2015--2017) \\
    \midrule
    Cyst                       & 275  & 109 \\
    HCC   & 214  &  79 \\
    Post-Treatment             & 205  &  79 \\
    Cirrhosis                  & 168  &  85 \\
    Steatosis                  & 154  &  93 \\
    Metastasis                 & 101  &  46 \\
    Hemangioma                 &  90  &  29 \\
    \midrule
    Total reports              & 773  & 325 \\
    \bottomrule
  \end{tabular}
\end{table}

\begin{figure*}[t]%
  \centering
  \includegraphics[width=0.9\linewidth,keepaspectratio]{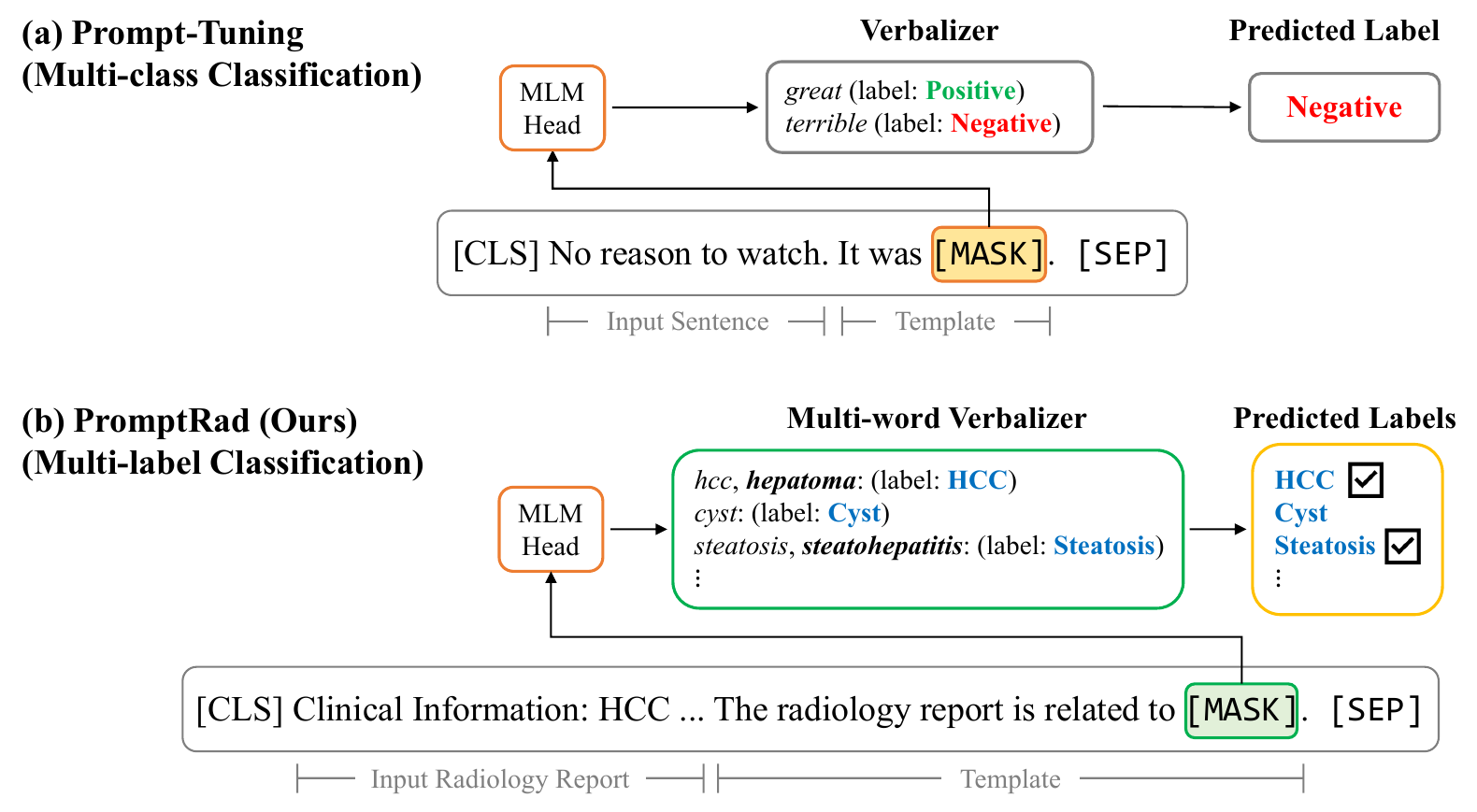}
  \caption{
    Differences in the approaches of (a) Prompt-tuning for multi-class classification on a general domain task \cite{gao-etal-2021-making,schick-schutze-2021-exploiting}. (b) PromptRad, our method for multi-label classification for radiology report labeling.
    Bold italic words in the verbalizer denote synonyms augmented from the UMLS Metathesaurus.
    MLM: masked language model \cite{devlin2019bert}.
  }
  \label{fig:main}
\end{figure*}

\section{Dataset}\label{sec:dataset}
This study was approved by the institutional review board of the participating medical center.
The dataset contains de-identified reports written in English, covering liver CT examinations from 2008 to 2017 (an example report is shown in Figure~\ref{fig:example_report}).
The reports were annotated by two senior radiologists for the presence of seven common cancer-related findings.
Each finding is labeled as positive (1) or negative (0); suspicious findings were treated as positive to avoid false negatives.

We split the data chronologically to ensure independence between training and test sets.
The 2008--2014 subset contains 773 reports and serves as the candidate training pool, while the 2015--2017 subset contains 325 reports and is used as the fixed test set, as summarized in Table~\ref{tab:dataset}.
To study low-resource scenarios, where only a small number of expert-annotated training examples are available,
for each run we randomly sample $K$ reports from the 773-report training pool using stratified sampling~\cite{10.1007/978-3-642-23808-6_10} to preserve the class distribution.
Our goal is to evaluate how effectively different methods learn under constrained annotation scenarios, rather than to assume that all 773 reports are unavailable.

\section{Method}
\subsection{Problem Definition}\label{method}
Given a radiology report $r$ and a label space $\mathcal{Y} = \{y_1, y_2, ..., y_n\}$ with $n$ distinct classes of findings, the report labeling task is to predict the presence of each finding $y_i \in \mathcal{Y}$ within $r$.
Multiple findings may be present in a single report.
We define the training set as $\mathcal{D}_\mathrm{train} = \{(r_a, \mathcal{Y}_a)\}_{a=1}^K$, where $K$ is the number of training examples and $\mathcal{Y}_a \subseteq \mathcal{Y}$ is the set of findings in report $r_a$.

\subsection{Standard Prompt-Tuning}\label{sec:prompt_tuning}
Prompt-tuning~\cite{pet2021,gao-etal-2021-making} fine-tunes PLMs without additional classification layers by (1) formatting the input as a cloze test with a textual prompt template containing a $\texttt{[MASK]}$ token, and (2) defining a verbalizer that maps each class to a word in the PLM vocabulary.
This allows fine-tuning with the MLM objective~\cite{devlin2019bert} without introducing new parameters.
We illustrate standard prompt-tuning in Figure~\ref{fig:main}~(a).

\begin{figure*}[t]%
    \centering
    \includegraphics[width=0.65\linewidth,keepaspectratio]{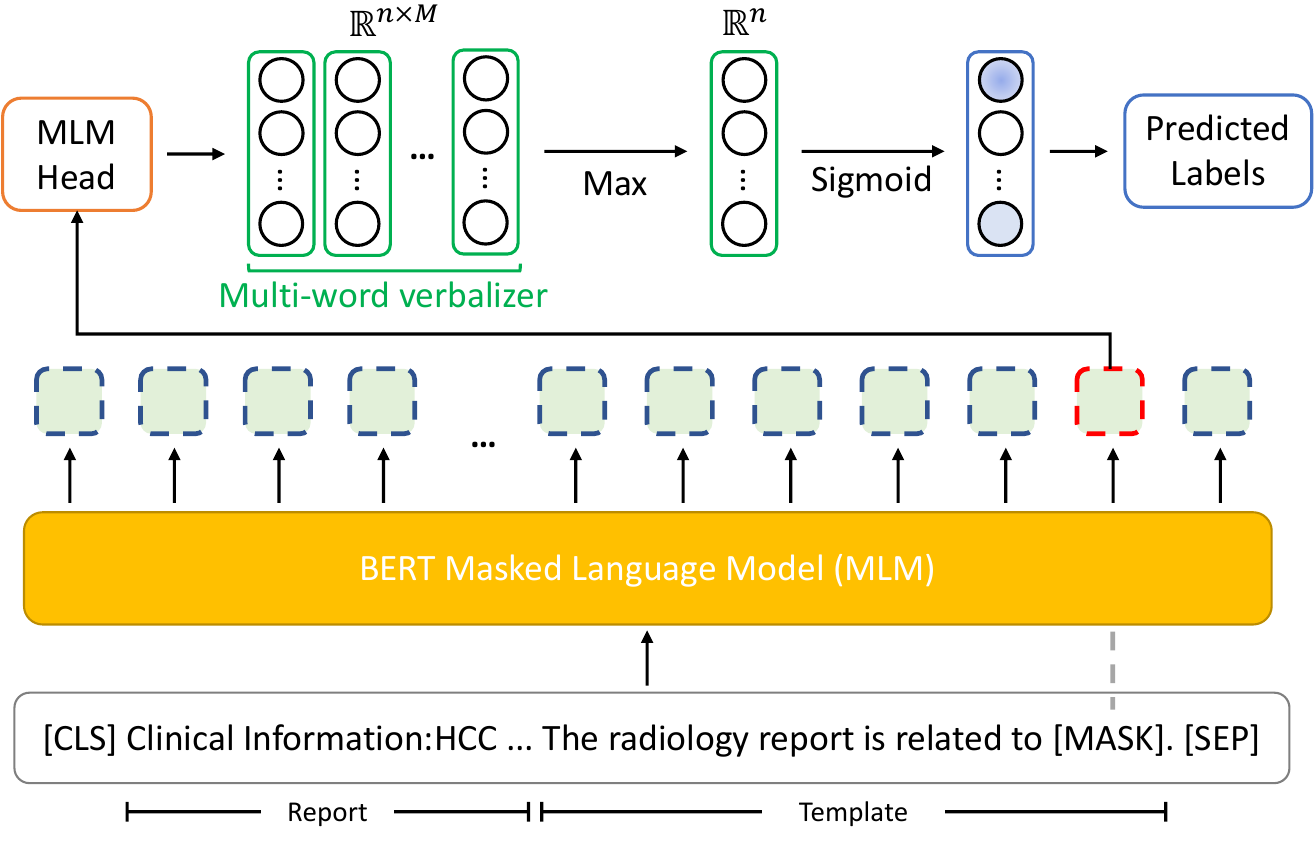}
    \caption{The workflow of the PromptRad report labeling system.}
    \label{fig:promptrad_flow}
\end{figure*}

\subsection{Multi-word Verbalizer of PromptRad}\label{sec:promptrad}
To identify the presence of each finding $y_i \in \mathcal{Y}$ in a report $r$ and transform report labeling into a masked language modeling task \cite{devlin2019bert,gao-etal-2021-making,schick-schutze-2021-exploiting}, we define a verbalizer $v: \mathcal{Y}\rightarrow V$ which maps a finding $y_i$ to a word from the vocabulary $V$ of a pre-trained MLM $\mathcal{M}$.
We first create a single-word verbalizer, where each finding is mapped to the word that most closely matches its class name (e.g., ``Post-treatment'' $\rightarrow$ ``posttreatment'').
Hence, if the model fills the word ``posttreatment'' in the $\texttt{[MASK]}$ position, the report will be classified as containing the ``Post-treatment'' finding.

In addition to the closest match, synonyms of the class names can also represent the findings.
To prevent the model from being biased towards a single mapping per category, we extend the verbalizer with synonyms from SNOMEDCT in the UMLS Metathesaurus~\cite{umls2004}.
For instance, both ``hcc'' and ``hepatoma'' serve as mappings for ``Hepatocellular Carcinoma,'' as shown in Figure~\ref{fig:main}~(b).
The complete mappings are listed in Appendix~\ref{subsec:verbalizer_mappings}.
This multi-word verbalizer can be easily adapted to other report labeling tasks by replacing the mappings with synonyms for the target categories.

\subsection{Training}\label{sec:promptrad_training}
As illustrated in Figure \ref{fig:promptrad_flow}, we append a prompt template to each report $r$ to form the input sequence $r_\textrm{p}$:

\vspace{4pt}
\noindent\hspace{1em}$r_\textrm{p}$ = \texttt{[CLS]} $r$ \text{The radiology report is related to} \texttt{[MASK]}\text{.} \texttt{[SEP]}
\vspace{4pt}

\noindent where the prompt template is manually designed.
Our objective is to determine the presence of each finding by the prediction of the $\texttt{[MASK]}$ token in $r_\textrm{p}$ using a pre-trained MLM $\mathcal{M}$.

During training, for each input $r_\textrm{p}$, we identify the presence of each finding $y_i \in \mathcal{Y}$ by recovering the $\texttt{[MASK]}$ token:
\begin{equation}
  z_i = f_\mathcal{M}(\texttt{[MASK]}, v(y_i) \mid r_\textrm{p})
\end{equation}
where $f_\mathcal{M}(\texttt{[MASK]}, v(y_i) \mid r_\textrm{p})$ denotes the logit score of token $v(y_i)$ at the \texttt{[MASK]} position produced by the MLM head of $\mathcal{M}$.
\noindent When multiple label-to-word mappings are available for a finding $y_i$, the one with the highest likelihood is chosen:
\begin{equation}\label{method:multiword}
  z_i=\mathrm{max}_{m=1}^{M_i}(f_\mathcal{M}(\texttt{[MASK]}, v(y_i, m)|r_\textrm{p})),
\end{equation}
\noindent where $M_i$ is the number of label-to-word mappings for $y_i$.
For example, for ``Hepatocellular Carcinoma'', $M_i$ is 2 with mappings ``hcc'' and ``hepatoma''.
Practically, as we show in Figure \ref{fig:promptrad_flow}, we create a matrix with the shape of $n\times M$ to facilitate the max operation in Equation \ref{method:multiword}, where $M$ is the maximum number of mappings among the categories.
We pad categories with fewer than $M$ mappings using dummy entries and apply a binary mask so that padded positions do not affect the maximum operation.
Then, we optimize $\mathcal{M}$ with the binary cross-entropy loss:
\begin{equation}
  \hat{y}_i = \mathrm{sigmoid}(z_i)
\end{equation}
\begin{equation}
  \mathcal{L} = -\frac{1}{n} \sum_{i=1}^{n}\left[ y_{i} \log(\hat{y}_i) + (1-y_{i}) \log(1-\hat{y}_i) \right].
\end{equation}

\noindent During inference, we assign a finding $y_i$ to a radiology report $r$ if the probability $\hat{y}_i$ exceeds a defined threshold $\tau$ for all the categories.

\subsection{Automatic Prompt Generation}\label{sec:autoT}
We argue that better performance can be achieved by training our model with different templates.
Instead of relying solely on manually designed templates, we explore automatic prompt generation following~\citet{gao-etal-2021-making}.

We take advantage of the generative capability of the pre-trained T5 \cite{T5} to automatically create various prompt templates and find the best one for our task.
Given a report $r$, the single-word verbalizer $v$, and a label $y_i$, we explore two formats: (1) report first: $[r]~[\mathbf{P}_1]~[v(y_i)]~[\mathbf{P}_2]$,
and (2) answer first: $[\mathbf{P}_1]~[v(y_i)]~[\mathbf{P}_2]~[r]$,
where $\mathbf{P}_1$ and $\mathbf{P}_2$ are T5-generated prompts placed before and after the label word.
Format 1 mirrors our manual template by placing the report before the label.
In addition, since the position of the $\texttt{[MASK]}$ token can vary in a sequence during the pre-training of BERT \cite{devlin2019bert}, we also explore Format 2, where the order is reversed.

\begin{table*}[t]
  \centering
  \caption{Performance comparison in F1-score (\%) using the independent test set.}
  \label{tab:promptrad_main}
  \setlength{\tabcolsep}{5.2pt}
  \begin{tabular}{@{}lccccccccc@{}}
    \toprule
    & \textbf{Cyst} & \textbf{HCC} & \textbf{Post-T} & \textbf{Cirr.} & \textbf{Stea.} & \textbf{Meta.} & \textbf{Hem.} & \textbf{Mac. F1} & \textbf{Mic. F1} \\
    \midrule
    GPT-4
      & 86.1 & 91.5 & 79.1 & 96.6 & 98.9 & 73.8 & 95.1 & 88.7 & 88.7 \\
    GPT-4 (ICL\textsuperscript{a})
      & 89.7 & 88.8 & 76.3 & 92.5 & 92.0 & 78.5 & 94.9 & 87.5 & 87.4 \\
    \midrule
    Label Match
      & 67.5 & 88.5 & 0.0 & 87.7 & 0.0 & 48.6 & \textbf{98.3} & 55.8 & 62.3 \\
    MetaMap
      & 77.6 & 86.9 & 48.6 & 53.5 & 95.7 & 27.5 & 84.6 & 67.8 & 69.1 \\
    NegBio
      & 77.6 & 86.9 & \textbf{81.4} & 82.7 & 95.7 & 27.5 & 84.6 & 76.6 & 79.2 \\
    \midrule
    PMB\textsuperscript{b}
      & 53.7\textsubscript{8.4} & 80.4\textsubscript{5.2} & 70.4\textsubscript{4.3} & 64.5\textsubscript{12.7} & 48.3\textsubscript{12.9} & 54.9\textsubscript{18.1} & 37.7\textsubscript{21.7} & 58.6\textsubscript{10.0} & 60.9\textsubscript{8.2} \\
    PMB\textsuperscript{b}+MM
      & 68.1\textsubscript{5.4} & 83.9\textsubscript{1.7} & 64.5\textsubscript{2.3} & 68.0\textsubscript{6.6} & 84.5\textsubscript{6.7} & 47.1\textsubscript{10.0} & 70.1\textsubscript{15.2} & 69.5\textsubscript{5.0} & 70.3\textsubscript{3.8} \\
    PMB\textsuperscript{b}+NB
      & 68.1\textsubscript{5.4} & 83.9\textsubscript{1.7} & 76.8\textsubscript{0.3} & 81.0\textsubscript{3.7} & 84.5\textsubscript{6.7} & 47.1\textsubscript{10.0} & 70.1\textsubscript{15.2} & 73.1\textsubscript{4.5} & 74.1\textsubscript{3.5} \\
    \textbf{PR}\textsuperscript{b}
      & 78.0\textsubscript{5.3} & 89.1\textsubscript{0.8} & 76.9\textsubscript{1.8} & 86.2\textsubscript{4.2} & 95.7\textsubscript{3.7} & 71.9\textsubscript{3.5} & 88.4\textsubscript{5.6} & 83.7\textsubscript{2.1} & 84.1\textsubscript{1.7} \\
    \textbf{PR+AutoT}\textsuperscript{b}
      & \textbf{89.5}\textsubscript{2.4} & \textbf{90.8}\textsubscript{1.0} & 78.4\textsubscript{3.9} & \textbf{91.0}\textsubscript{3.4} & \textbf{97.3}\textsubscript{0.9} & \textbf{84.7}\textsubscript{1.2} & 92.4\textsubscript{2.5} & \textbf{89.2}\textsubscript{1.0} & \textbf{89.4}\textsubscript{1.0} \\
    \bottomrule
  \end{tabular}

\vspace{4pt}
\raggedright\footnotesize
Abbreviations: Cirr.: Cirrhosis; Stea.: Steatosis; Meta.: Metastasis; Hem.: Hemangioma;
Mac.: Macro average; Mic.: Micro average; Post-T: Post-Treatment;
PMB: PubMedBERT; MM: MetaMap; NB: NegBio; PR: \textbf{P}rompt\textbf{R}ad (ours).\\
\textsuperscript{a}~In-Context Learning with three random samples from the training set.
\textsuperscript{b}~Average from five runs; subscripts: standard deviation.
\end{table*}

For each report with multiple labels, we duplicate the input for each label and use T5 with beam search to generate 100 candidate templates (50 per format), ranked by the joint log-likelihood over all training examples~\cite{gao-etal-2021-making}.
We train PromptRad with each candidate and select the best-performing template (Table~\ref{tab:autoT} in Appendix~\ref{subsec:auto_templates}).

\section{Experiments}
We evaluate PromptRad against a range of baselines on the test set.
Unless otherwise specified, all training-based methods (e.g., PubMedBERT and PromptRad) use 32 labeled reports ($K{=}32$) as training data.
Detailed experimental settings, baseline descriptions, and GPT-4 prompts are provided in Appendix~\ref{subsec:imp}.
\subsection{Performance Comparison}\label{result:main}
Table \ref{tab:promptrad_main} shows that the proposed PromptRad method outperforms all the baselines for most of the categories using the manual template, except for GPT-4 \cite{openai2023gpt}.
In addition, Table \ref{tab:promptrad_main} also shows that PromptRad+AutoT benefits from the proposed automatic template generation method and slightly outperforms GPT-4 in terms of average scores.
This result suggests that fine-tuning BERT using appropriate prompts provides a competitive and lightweight alternative to API-based large language models for low-resource report labeling.

For the category of ``Hemangioma", ``Label Match" can almost perfectly identify the positive findings, showing that the descriptions for ``Hemangioma" can be easily identified.
The reason that our approaches perform worse than ``Label Match" in the ``Hemangioma" category is that the number of training examples for ``Hemangioma" is relatively small\footnote{During pre-processing, we randomly sampled 32 examples while keeping the distribution of the sampled data the same as that of the original training set, as we mentioned in Section \ref{sec:dataset}. Thus, there are only 2-3 samples labeled as ``Hemangioma'' each time.}, compared to the other categories.
We will show the performance of PromptRad with more training examples in Section \ref{sec:sizes}.
We note that ``Label Match'' scores zero on ``Post-Treatment'' and ``Steatosis'' because these categories are expressed through synonyms (e.g., RFA, TACE, fatty liver) that do not match the category names directly.

\subsection{Case Study}\label{result:examples}
To confirm that the proposed method is capable of handling diverse descriptions for the target labels, we perform a case study and list the examples with the corresponding labels in Table \ref{tab:examples}.
In this experiment, we compare the performance of our method with NegBio \cite{peng2018negbio} as representative of the dictionary-based methods, since NegBio outperforms MetaMap \cite{aronson2001effective} in our experiment (Table \ref{tab:promptrad_main}).
Such cases are challenging for dictionary-based systems because the target label names may appear in the text even when the corresponding findings are negated or absent.

In Table \ref{tab:examples}, we observe that most of the positive findings can be correctly identified by both approaches when the label names are explicitly mentioned in the reports.
However, NegBio usually fails to negate the findings when the label names show up in reports with negative descriptions, such as ``No ... nor HCC in both lobes liver'' and ``No imaging evidence of cirrhosis ...'' in the first two examples of Table \ref{tab:examples}.
Additionally, NegBio can be misled by common negation abbreviations in radiology reports, such as ``R/O'' (rule out) in the third report of Table \ref{tab:examples}, while learning-based methods, such as PromptRad, can learn to recognize these abbreviations from the training data.

This experiment demonstrates the potential of the proposed method in handling negation cases.
To further validate the effectiveness of the proposed method for handling negation cases in radiology reports, we next perform a quantitative experiment.

\begin{table}[htbp]
  \centering
  \caption{
    Case study comparing PromptRad+AutoT with NegBio on reports containing negated findings. Bold italic labels indicate gold annotations for each sentence.
  }
  \label{tab:examples}
  \small
  \begin{tabular}{@{}lp{0.8\columnwidth}@{}}
    \toprule
    & Segments (\textit{Gold Label}) and Predicted Positive Labels \\
    \midrule
    Report 1
      & A tiny hepatic cyst in S7/8> Patency of the SMV \textbf{(\textit{Positive: Cyst})} \\
      & No obvious hypervascular nodule nor HCC in both lobes liver. \textbf{(\textit{Negative: HCC})} \\
      & Mild liver cirrhosis and portal hypertension. \textbf{(\textit{Positive: Cirrhosis})} \\
    \addlinespace
      & \textbf{NegBio}: Cyst, HCC, Cirrhosis \\
      & \textbf{PromptRad+AutoT}: Cyst, Cirrhosis \\
    \midrule
    Report 2
      & No imaging evidence of cirrhosis of liver. \textbf{(\textit{Negative: Cirrhosis})} \\
      & Two small 0.6-cm and 1.4-cm densely packed lipiodol puddles in S7 without
        identifiable viable tumor, suggestive of good response to previous TACE without viability. \textbf{(\textit{Positive: HCC, Post-Treatment})} \\
      & Multiple hepatic cysts are noted. \textbf{(\textit{Positive: Cyst})} \\
    \addlinespace
      & \textbf{NegBio}: Cyst, HCC, Post-Treatment, Cirrhosis \\
      & \textbf{PromptRad+AutoT}: Cyst, HCC, Post-Treatment \\
    \midrule
    Report 3
      & Remarkable fatty liver with GB stones; \textbf{(\textit{Positive: Steatosis})} \\
      & R/O metastasis in the anterior abdominal wall. \textbf{(\textit{Negative: Metastasis})} \\
    \addlinespace
      & \textbf{NegBio}: Steatosis, Metastasis \\
      & \textbf{PromptRad+AutoT}: Steatosis \\
    \bottomrule
  \end{tabular}
\end{table}

\subsection{Effectiveness of Negation Handling}\label{result:negation}
To quantitatively assess the effectiveness of our proposed method in handling negation cases in radiology reports,
we collect the reports from the test set that explicitly mentioned the label names, but were annotated as negative findings by our radiologists.
We excluded the ``Cyst", ``Post-Treatment", ``Steatosis", and ``Hemangioma" categories, as reports mentioning these label names usually indicate positive findings of the corresponding categories.

\begin{figure}[htbp]%
  \centering
  \includegraphics[width=1.0\linewidth,keepaspectratio]{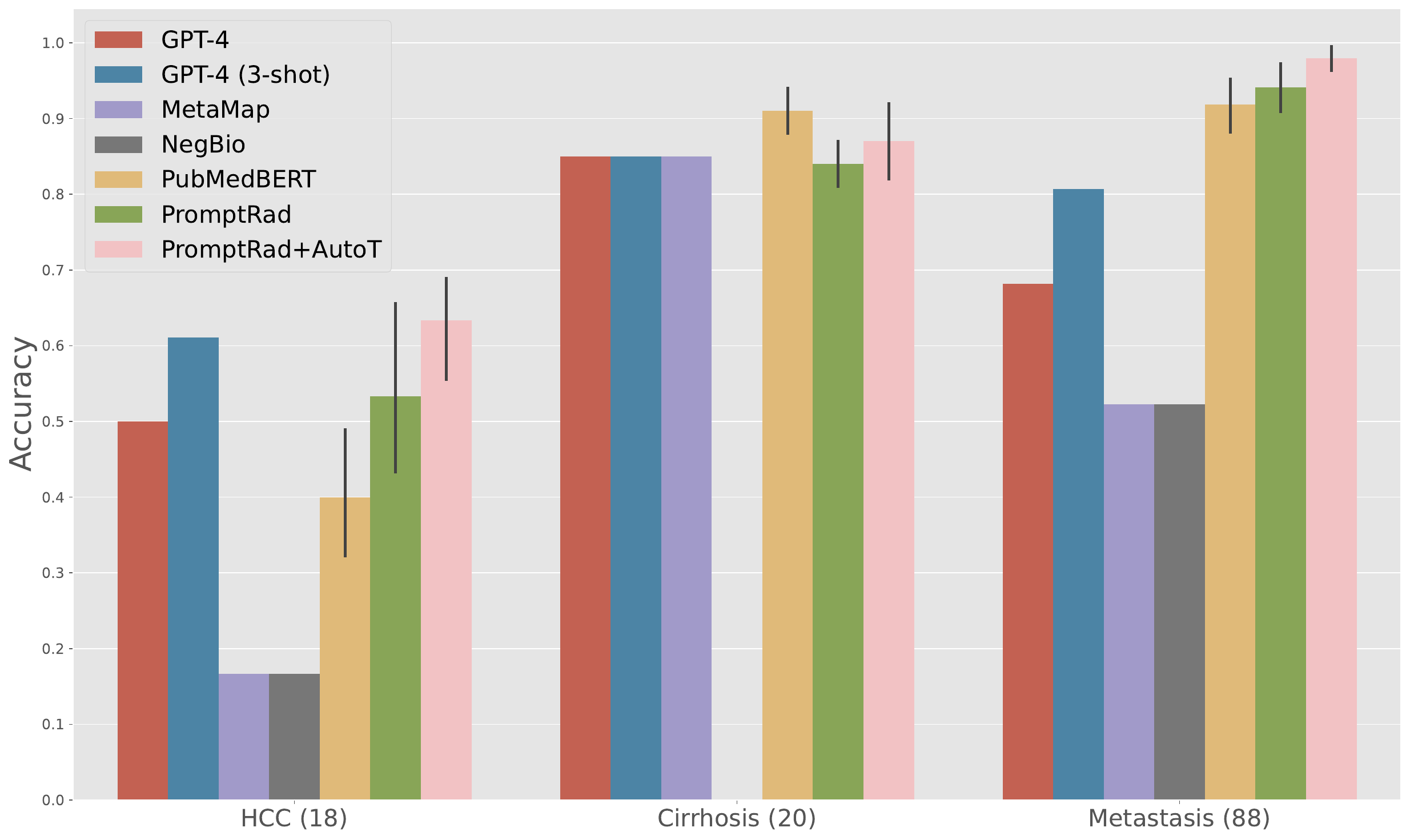}
  \caption{
    Accuracy on negation cases: reports that mention a finding name but are annotated as negative.
    Values in parentheses indicate the number of cases per category.
  }
  \label{fig:negation}
\end{figure}

Figure \ref{fig:negation} shows that PromptRad+AutoT (ours) is better at handling negation cases than GPT-4 \cite{openai2023gpt}, NegBio \cite{peng2018negbio}, and MetaMap \cite{aronson2001effective} for the negated reports in the three categories.
Additionally, our manual-template approach (PromptRad) remains competitive compared to the other baselines.
To our surprise, the performance of NegBio on ``Cirrhosis" is unsatisfactory.
This is due to the fact that NegBio relies heavily on natural language rules of dependency parsing and takes inputs with complete sentence structures.
However, the descriptions relevant to ``Cirrhosis" in our dataset are usually short (e.g., \textit{No liver cirrhosis.}), which makes it difficult for NegBio to identify the negation cases.
Besides, though our methods are slightly inferior to PubMedBERT \cite{pubmedbert} for the reports with ``Cirrhosis", they are still much better at handling negation cases than PubMedBERT for the reports with ``HCC" and ``Metastasis".
Based on the overall performance in Figure \ref{fig:negation}, we can still conclude that our method is a better choice for handling negation cases in radiology reports than the other baselines.

\begin{figure*}[htbp]
  \centering
  \includegraphics[width=0.99\linewidth,keepaspectratio]{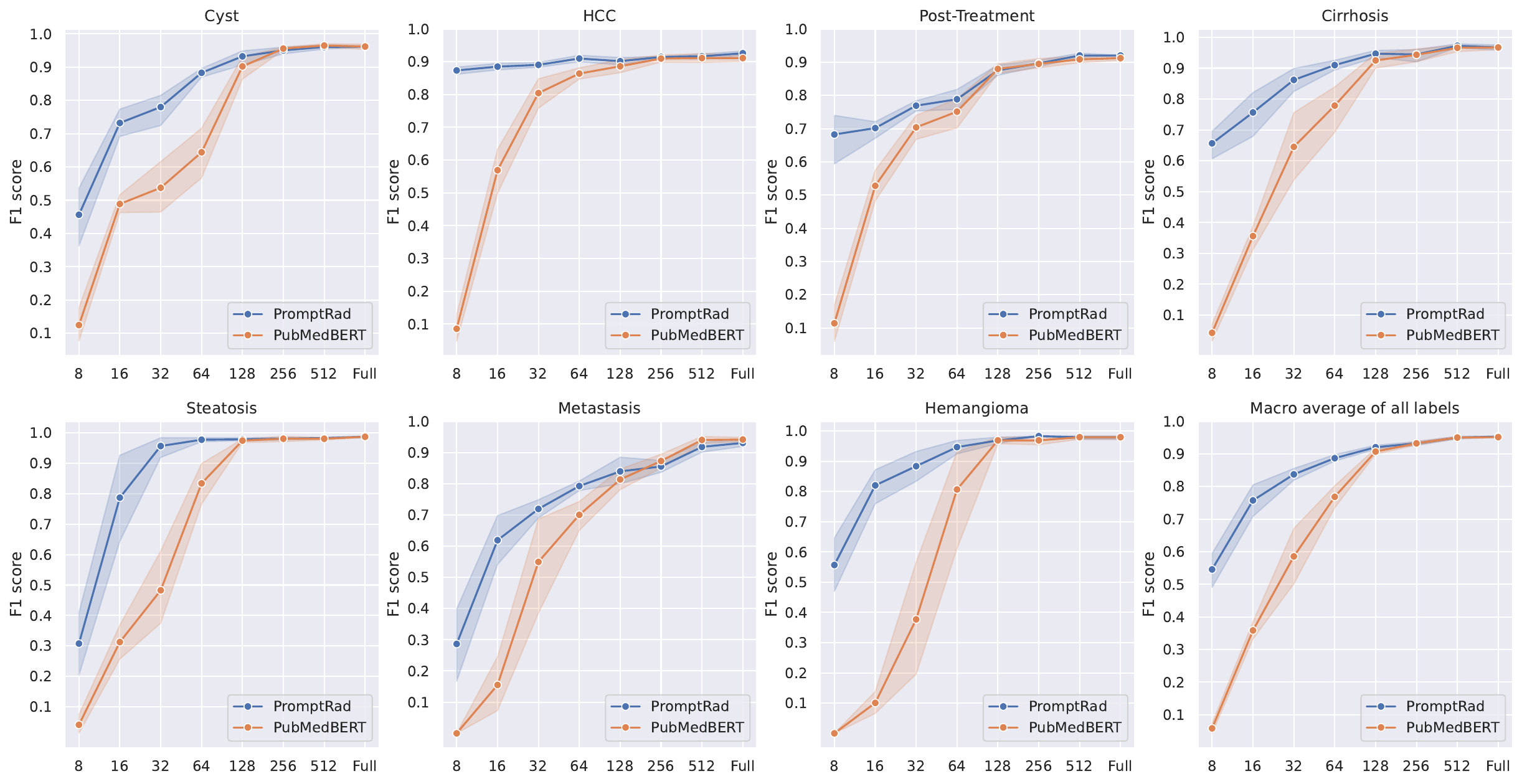}
  \caption{Comparison in F1-score for using different numbers of reports for training. We ran experiments five times for each training size, and for a training size of 8, we repeated the experiment ten times. ``Full" indicates the use of the full training set.}
  \label{fig:sizes}
\end{figure*}

\subsection{Performance Comparison for Different Training Sizes}\label{sec:sizes}
In the previous sections, we studied the performance of the proposed methods for using only 32 labeled reports for training.
In this section, we investigate the impact of the training size on the performance of the report labeling task.
We randomly sampled different numbers of reports from the training set and evaluated the performance of PromptRad and PubMedBERT on the test set.
Figure~\ref{fig:sizes} shows that both methods exhibit a similar trend in performance improvement with the increase of training sizes, and both methods achieve more than 90\% F1-score in macro average when the training size is larger than 128 samples.
In addition, the proposed method outperforms PubMedBERT in almost all categories with training sizes smaller than 128 samples, showing that our method is effective under data scarcity scenarios.

\subsection{Ablation Study for the Verbalizer}\label{mprt_app:verb}
Since PromptRad and PromptRad+AutoT use the multi-word verbalizer,
to confirm the effectiveness of the multi-word verbalizer,
we provide a performance comparison for the two types of verbalizers in Table \ref{tab:verbalizer_scores}.
We find that our methods using the multi-word verbalizer outperform the ones using the single-word verbalizer on average across all seven categories.
Notably, the multi-word verbalizer also reduces variance across runs (e.g., Cirrhosis std.\ drops from 12.6 to 4.2 for PromptRad), suggesting that multiple mappings provide more stable category representations during few-shot training.
Therefore, we conclude that the inclusion of multiple label-to-word mappings is beneficial for the radiology report labeling task.

We further note that, according to the complete mappings in Appendix~\ref{subsec:verbalizer_mappings}, the two verbalizers differ only for ``Hepatocellular Carcinoma'' and ``Steatosis''.
Since PromptRad is trained with shared PubMedBERT parameters and a shared MLM head, these mapping changes may indirectly affect other categories through the joint multi-label objective.
Thus, differences for categories with identical mappings likely reflect shared-parameter effects.
For other report labeling tasks, the multi-word verbalizer can be easily extended by adding synonyms for additional target categories when available.

\begin{table}[t]
  \centering
  \caption{Performance comparison in F1-score for PromptRad and PromptRad+AutoT using the single-word and multi-word verbalizers.}
  \label{tab:verbalizer_scores}
  \small
  \begin{tabular}{@{}lcccc@{}}
    \toprule
    & \multicolumn{2}{c}{PromptRad} & \multicolumn{2}{c}{PromptRad+AutoT} \\
    \cmidrule(lr){2-3}\cmidrule(lr){4-5}
    Labels & Single & Multi & Single & Multi \\
    \midrule
    Cyst           & 79.1\textsubscript{4.2}  & 78.0\textsubscript{5.3}  & 89.4\textsubscript{3.2} & \textbf{89.5}\textsubscript{2.4} \\
    HCC            & 89.4\textsubscript{0.6}  & 89.1\textsubscript{0.8}  & 89.9\textsubscript{0.6} & \textbf{90.8}\textsubscript{1.0} \\
    Post-Treatment & 74.4\textsubscript{2.4}  & \textbf{76.9}\textsubscript{1.8} & \textbf{80.2}\textsubscript{3.7} & 78.4\textsubscript{3.9} \\
    Cirrhosis      & 84.2\textsubscript{12.6} & \textbf{86.2}\textsubscript{4.2} & 90.3\textsubscript{2.5} & \textbf{91.0}\textsubscript{3.4} \\
    Steatosis      & 96.3\textsubscript{2.2}  & 95.7\textsubscript{3.7}  & 95.7\textsubscript{2.2} & \textbf{97.3}\textsubscript{0.9} \\
    Metastasis     & 74.0\textsubscript{5.8}  & 71.9\textsubscript{3.5}  & 83.2\textsubscript{1.7} & \textbf{84.7}\textsubscript{1.2} \\
    Hemangioma     & 83.6\textsubscript{5.8}  & \textbf{88.4}\textsubscript{5.6} & 89.1\textsubscript{7.8} & \textbf{92.4}\textsubscript{2.5} \\
    \midrule
    Macro Avg.     & 83.0\textsubscript{3.0}  & \textbf{83.7}\textsubscript{2.1} & 88.2\textsubscript{1.6} & \textbf{89.2}\textsubscript{1.0} \\
    Micro Avg.     & 84.1\textsubscript{2.7}  & \textbf{84.1}\textsubscript{1.7} & 88.9\textsubscript{1.4} & \textbf{89.4}\textsubscript{1.0} \\
    \bottomrule
  \end{tabular}

  \vspace{4pt}
  \raggedright\footnotesize
  Scores were averaged from five runs. Subscripts: standard deviation.
\end{table}

\section{Discussion}
\subsection{Why PromptRad Works}
PromptRad outperforms standard BERT fine-tuning with PubMedBERT~\cite{pubmedbert} across all categories when only 32 labeled reports are available (Table~\ref{tab:promptrad_main}), and this advantage persists for training sizes below 128 (Figure~\ref{fig:sizes}).
We attribute this to the alignment between prompt-tuning and masked language model pre-training.
Unlike standard fine-tuning, which introduces a randomly initialized task-specific classification layer, PromptRad introduces no additional task-specific classification parameters. It reuses the pre-trained MLM head and predicts labels through verbalized clinical concepts, preserving closer alignment with the masked language modeling objective used during pre-training.

\subsection{Advantages over Rule-based and API-based Alternatives}
Compared to dictionary-based systems such as MetaMap~\cite{aronson2001effective} and NegBio~\cite{peng2018negbio}, PromptRad learns from a small number of labeled reports and better handles diverse clinical descriptions and negation patterns (Figure~\ref{fig:negation}), which rule-based systems often miss due to reliance on predefined patterns.

PromptRad also achieves competitive performance with GPT-4~\cite{openai2023gpt} under our low-resource setting while using a lightweight PubMedBERT backbone.
Beyond efficiency, a locally deployable model avoids sending sensitive clinical reports to external APIs, which is an important consideration for real-world clinical deployment. We therefore view PromptRad not as a replacement for general-purpose large language models, but as a practical alternative when data privacy, deployment cost, and limited annotation resources are central constraints.

\subsection{Comparison with Local LLMs}
A natural alternative to API-based LLMs is deploying open-source LLMs locally (e.g., Llama~3 \cite{llama3}, Qwen \cite{yang2025qwen3}).
While this addresses privacy concerns, it requires substantial GPU memory (typically 16GB+ VRAM even for 7--8B models) that may be unavailable in smaller clinical institutions.
In contrast, PromptRad uses a 110M-parameter backbone that can be deployed on consumer-grade hardware, including CPU-only inference.
This makes PromptRad practical for resource-constrained clinical environments where both privacy and deployment cost are concerns.

\section{Conclusion}
In this paper, we introduce PromptRad, a knowledge-enhanced prompt-tuning approach for multi-label radiology report labeling under low-resource settings.
Using a UMLS-based multi-word verbalizer and only 32 labeled reports, PromptRad outperforms dictionary-based and fine-tuning baselines, achieves competitive performance with GPT-4, and handles complex negation patterns more effectively than existing methods.
As a lightweight, locally deployable model, PromptRad offers a practical solution for clinical report labeling where data privacy and annotation costs are central concerns, and can further support applications such as benign lesion retrieval and large-scale radiograph annotation.

\section*{Limitations}
This study is based on liver CT reports from a single medical center, which may limit the generalizability of PromptRad to reports from other institutions, imaging domains, or reporting styles.
In addition, our experiments focus on English reports and seven predefined liver-related findings, so further validation is needed to assess whether the method transfers well to broader clinical settings.
Furthermore, the number of reports included in the negation-focused analysis is relatively small, as indicated by the counts in Figure~\ref{fig:negation}. Larger annotated datasets are needed to better characterize model robustness on challenging negation cases.
Finally, the construction of the multi-word verbalizer also depends on an external knowledge base such as the UMLS Metathesaurus.
In settings where suitable terminology resources are unavailable, the applicability of this design may be reduced.

\section*{Acknowledgments}
This work was supported by the National Science and Technology Council in Taiwan, under grants NSTC 114-2223-E-007-011 and NSTC 114-2222-E-182-001-MY2.
We would like to express our deepest gratitude to the reviewers for their thoughtful comments and valuable feedback.
We would also like to thank Szu-Tung Lin for his assistance in developing the data annotation platform.


\bibliography{custom}

\appendix

\section{Appendix}
\label{sec:appendix}
\subsection{Verbalizer Mappings}\label{subsec:verbalizer_mappings}
As shown in Table~\ref{tab:verbalizer}, the differences between the single-word and multi-word verbalizers fall into the categories of ``Hepatocellular Carcinoma'' and ``Steatosis'', as there are no additional synonyms for the remaining categories.

\subsection{Automatically Generated Templates}\label{subsec:auto_templates}
We list the automatically generated templates with the highest scores in Table~\ref{tab:autoT}.
We observe that the T5 model \cite{phan2021scifive} tends to generate short but concise prompts, such as ``Hepatic $\texttt{[MASK]}$'' or ``Liver $\texttt{[MASK]}$'', while the manual template (also listed in Table~\ref{tab:autoT}) is more descriptive.

\begin{table*}[t]
  \centering
  \caption{
    The label-to-word mappings (verbalizer) we used for PromptRad and PromptRad+AutoT.
  }
  \label{tab:verbalizer}
  \begin{tabular}{lll}
  \toprule
    \textbf{Category} & \textbf{Single-Word Verbalizer} & \textbf{Multi-Word Verbalizer}  \\
    \midrule
    Cyst              & cyst                            & cyst                            \\
    Hepatocellular Carcinoma  &  hcc            & hcc, \textbf{hepatoma} \\
    Cirrhosis         & cirrhosis                       & cirrhosis                       \\
    Post-Treatment    & posttreatment                   & posttreatment                   \\
    Steatosis         & steatosis                       & steatosis, \textbf{steatohepatitis}  \\
   Metastasis        & metastasis                      & metastasis                      \\
    Hemangioma        & hemangioma                      & hemangioma                      \\
  \bottomrule
  \end{tabular}
\end{table*}

\begin{table*}[t]
  \caption{
    The top-5 templates from automatic template generation along with the manual template for comparison.
    ``Manual'' was used in PromptRad, while ``AutoT$^{1}$'' was used in PromptRad+AutoT.
  }
  \label{tab:autoT}
  \centering
  \renewcommand{\arraystretch}{1.2}
  \begin{tabular}{ll}
      \toprule
      \textbf{Template} & \textbf{Format} \\
      \midrule
      Manual   & [\textit{Report}] The liver radiology report is related to $\texttt{[MASK]}$\text{.} \\
      \midrule
      AutoT$^{1}$    & Hepatic $\texttt{[MASK]}$: [\textit{Report}] \\
      AutoT$^{2}$    & Liver $\texttt{[MASK]}$: [\textit{Report}] \\
      AutoT$^{3}$    & [\textit{Report}] Hepatic $\texttt{[MASK]}$. \\
      AutoT$^{4}$    & Abdominal $\texttt{[MASK]}$. [\textit{Report}] \\
      AutoT$^{5}$    & Liver $\texttt{[MASK]}$ in [\textit{Report}] \\
      \bottomrule
  \end{tabular}
\end{table*}

\section{Implementation Details}\label{subsec:imp}
We implement our model using the HuggingFace Transformers \cite{wolf-etal-2020-transformers} (version: 4.12.2) and the PyTorch \cite{pytorch} (version: 1.10.0) libraries.
We used NVIDIA GeForce GTX 1070 and RTX 2080 Ti GPUs for training our model and finding templates with automatic prompt generation under the CUDA version of 10.1.

To ensure fair comparisons,
all experiments for PromptRad and PubMedBERT use the model checkpoint of the pre-trained PubMedBERT-base\footnote{\url{https://huggingface.co/microsoft/BiomedNLP-PubMedBERT-base-uncased-abstract}} \cite{pubmedbert} for initialization,
and we use SciFive-base\footnote{\url{https://huggingface.co/razent/SciFive-base-Pubmed_PMC}} \cite{phan2021scifive} for automatic prompt generation.
For optimizing the models, we use AdamW \cite{adamw} as the optimizer to train the models.
All experiments are reported with 32 labeled reports as training data, unless otherwise specified.
To simulate the real-world scenario of only limited labeled data available, we do not use a development set for hyperparameter tuning.
Instead, we choose the best configuration according to the training loss.
We use grid search with the following ranges:
\begin{itemize}
  \item Learning rate: $[2e-5, 3e-5, 5e-5]$
  \item Batch size: $[2, 4, 8]$
  \item Threshold $\tau$: $[0.2, 0.3, 0.4, 0.5]$
  \item Ratio of training steps for learning rate warmup: $[0.0, 0.1]$
\end{itemize}

\subsection{Baseline Methods}\label{sec:baselines}
This section describes the baseline methods we used for comparison in our experiments.
\subsubsection{GPT-4}
We used the OpenAI API\footnote{\url{https://openai.com/blog/openai-api}} to test GPT-4 on report labeling, with the model name\footnote{\url{https://platform.openai.com/docs/models}} of ``gpt-4".
Our input instructions can be found in Table \ref{tab:openai}, where they are displayed as both the system message and the user message.
For each liver CT report, we first replaced \textbf{[REPORT]} in the user message with the actual report text.
The modified user message was then processed by GPT-4 to produce the corresponding output assistant message.

\subsubsection{GPT-4 (ICL)}
The use of in-context learning (ICL) \cite{NEURIPS2020_1457c0d6,openai2023gpt} helps large language models achieve better performance by providing a few labeled examples in an input prompt.
In this study, we also compare our approach with GPT-4 (ICL).
We provide three random examples (3-shot) from the training set in the user message (Table \ref{tab:openai}) to the GPT-4 model to see if the performance can be improved through in-context learning.

\subsubsection{Label Match}
We create a straightforward rule-based baseline that identifies positive findings when a report contains the corresponding label names.
Our matching rules also account for plural forms of label names.
For instance, an occurrence of ``metastases" in a report is also recognized as a positive finding for ``metastasis".
This simple baseline can also be used to observe the difficulties of each category.

\subsubsection{MetaMap}
MetaMap \cite{aronson2001effective} is a widely used dictionary-based tool for extracting concepts from biomedical text by mapping text to the UMLS Metathesaurus \cite{umls2004}.
Due to its vast repository of biomedical knowledge and its integration of both hierarchical and non-hierarchical relationships using semantic types and relations, MetaMap stands as a fundamental and effective benchmark for report labeling tasks \cite{Wang_2017_CVPR}.
We use the 2020 version of MetaMap with pymetamap\footnote{\url{https://github.com/AnthonyMRios/pymetamap}}, an open source MetaMap wrapper, to extract the concepts for each report in our test set.

For each report, we obtain a positive finding if MetaMap returns one of the corresponding UMLS CUIs (Concept Unique Identifiers) of each category.
The mappings between CUIs and categories that we used for MetaMap to retrieve the positive findings in our task are provided in Table \ref{tab:matamap}.

\subsubsection{NegBio}
NegBio \cite{peng2018negbio} is a dictionary-based tool for negation and uncertainty detection in clinical text.
This approach should work in conjunction with either CheXpert \cite{irvin2019chexpert} or MetaMap \cite{aronson2001effective}.
We use the official implementation of NegBio\footnote{\url{https://github.com/ncbi-nlp/NegBio}} to extract the concepts for each report in our test set using the same CUIs in Table \ref{tab:matamap}.

\subsubsection{PubMedBERT}
We fine-tune the pre-trained PubMedBERT model \cite{pubmedbert} for multi-label classification using the HuggingFace API \cite{wolf-etal-2020-transformers}.
This baseline serves as a representative of the standard fine-tuning approach for PLMs.
Similar to the proposed method, we set a threshold $\tau$ to determine the positive findings for each label.
We perform hyperparameter tuning for PubMedBERT with the same hyperparameter ranges as the proposed method, as described in Section \ref{subsec:imp}.

\subsubsection{PubMedBERT+MM}
We create an ensemble of PubMedBERT \cite{pubmedbert} and MetaMap (MM) \cite{aronson2001effective} by
combining the predicted positive findings from PubMedBERT and MetaMap, denoted ``PubMedBERT+MM".
This baseline serves as a hybrid model that combines the strengths of both expert knowledge from MetaMap and the machine learning ability of PubMedBERT.

\subsubsection{PubMedBERT+NB}
In addition to PubMedBERT+MM, we also create ``PubMedBERT+NB", an ensemble of PubMedBERT \cite{pubmedbert} and NegBio (NB) \cite{peng2018negbio}.
As with PubMedBERT+MM, PubMedBERT+NB combines the predicted positive findings from PubMedBERT and NegBio to take advantage of both methods.

\subsection{GPT-4 Prompt Design}\label{subsec:gpt4_prompt}
Table~\ref{tab:openai} shows the input message we used to query GPT-4 via the OpenAI API for report labeling. The system message assigns the role of a radiologist, and the user message provides the classification task along with the report text. For GPT-4 (ICL), we additionally include three randomly sampled labeled examples from the training set in the user message.

\begin{table*}[t]
  \centering
  \caption{Input message and an example output of GPT-4 using the OpenAI API.}
  \label{tab:openai}
  \small
  \begin{tabular}{@{}m{2cm}m{13cm}@{}}
    \toprule
    \textbf{Role} & \textbf{Content} \\
    \midrule
    System &
      You are a professional radiologist who knows computed tomography (CT)
      very much. You can classify the CT report for the liver features or
      symptoms. \\
    \midrule
    User &
      Now you are going to perform a multi-label classification for a text
      report of liver computed tomography (CT). Given the potential
      categorized features \{`cyst': 0, `HCC': 1, `post-treatment': 2,
      `cirrhosis': 3, `steatosis': 4, `metastasis': 5, `hemangioma': 6\},
      please read the following liver computed tomography report of a
      patient. If the report is positive with a feature, please return the
      corresponding value. Liver computed tomography: \textbf{[REPORT]} \\[6pt]
    \midrule
    Assistant\newline Response\newline(Example) &
      1. HCC (Hepatocellular carcinoma)\newline
      3. Cirrhosis \\
    \bottomrule
  \end{tabular}
\end{table*}

\subsection{UMLS Concept Identifiers}\label{subsec:umls_cuis}

Table~\ref{tab:matamap} lists the UMLS Concept Unique Identifiers (CUIs) used with MetaMap and NegBio to identify positive findings for each category. Since ``Post-Treatment'' encompasses multiple treatment procedures, we consider a report positive for this category if any of the corresponding CUIs is detected.

\begin{table}[t]
  \centering
  \caption{
    Concept Unique Identifiers (CUIs) used with MetaMap \cite{aronson2001effective} and NegBio \cite{peng2018negbio} for our task.
    RFA: Radiofrequency Ablation; TACE: Transarterial Chemoembolization.
  }
  \label{tab:matamap}
  \small
  \begin{tabular}{@{}lp{0.66\columnwidth}@{}}
    \toprule
    \textbf{Category}               & \textbf{CUI Term (CUI)}                    \\
    \midrule
    Cyst                            & Liver cyst (C0267834)                      \\
    HCC                             & Liver carcinoma (C2239176)                 \\
    \multirow{4}{*}{Post-Treatment} & RFA (C0850292)         \\
                                    & TACE (C3539919) \\
                                    & Embolization, Therapeutic (C0013931)       \\
                                    & Lobectomy (C0023928)                       \\
    Cirrhosis                       & Liver Cirrhosis (C0023890)                 \\
    Steatosis                       & Fatty Liver (C0015695)                     \\
    Metastasis                      & Metastasis (C4255448)                      \\
    Hemangioma                      & Hemangioma (C0018916)                      \\
    \bottomrule
  \end{tabular}
\end{table}

\subsection{Evaluation Metric}\label{sec:evaluation}
We use the F1-score as the evaluation metric for the seven categories in our experiments.
We also include the micro-averaged F1-score and macro-averaged F1-score for the overall performance of each model.
All the experiments use the same test set mentioned in Section \ref{sec:dataset} for evaluations.
For the training-based methods, including PubMedBERT \cite{pubmedbert} and our approaches, we report the average scores and standard deviations of 5 runs with different random seeds.
For the experiment with only eight training examples, we report the results of 10 runs with different random seeds.
This is because the performance of the training-based methods can be more unstable when the training size is smaller, and we want to provide a more comprehensive evaluation for this experiment.

\end{document}